\documentclass{article}
\pdfoutput=1
\usepackage{arxiv}

\usepackage[utf8]{inputenc} 
\usepackage[T1]{fontenc}    
\usepackage{hyperref}       
\usepackage{url}            
\usepackage{booktabs}       
\usepackage{amsfonts}       
\usepackage{nicefrac}       
\usepackage{microtype}      
\usepackage{xcolor}         
\usepackage{graphicx}
\usepackage{amsmath}

\title{Generalization ability and Vulnerabilities to adversarial perturbations: Two sides of the same coin}

\author{%
  Jung H. Lee \\
  Pacific Northwest National Laboratory\\
  Seattle, WA \\
  \texttt{jung.lee@pnnl.gov} \\
  \And
  Sujith Vijayan \\
  School of Neuroscience\\
  Virginia Tech\\
  Blacksburg, VA \\
  \texttt{neuron99@vt.edu} \\
}

\begin{document}

\maketitle

\begin{abstract}
Deep neural networks (DNNs), the agents of deep learning (DL), require a massive number of parallel/sequential operations, which makes it difficult to comprehend them and impedes proper diagnosis. Without better knowledge of DNNs’ internal process, deploying DNNs in high-stakes domains may lead to catastrophic failures. Therefore, to build more reliable DNNs/DL, it is imperative that we gain insights into their underlying decision-making process. Here, we use the self-organizing map (SOM) to analyze DL models’ internal codes associated with DNNs’ decision-making. Our analyses suggest that shallow layers close to the input layer map onto homogeneous codes and that deep layers close to the output layer transform these homogeneous codes in shallow layers to diverse codes. We also found evidence indicating that homogeneous codes may underlie DNNs' vulnerabilities to adversarial perturbations. 
\end{abstract}

\section{Introduction}
Deep learning (DL) can train DL models (deep neural networks, DNNs) with examples to perform a wide range of real-world problems \cite{Lecun2015, deep_review1, deep_review2}. This may remove the need for human experts’ instructions, but DL models depend on a large number of linear and nonlinear operations, making their operations incomprehensible. Without more comprehensive knowledge of their operations, we cannot conduct timely predictions of failures or perform proper diagnosis, and without proper diagnosis, we cannot safely deploy DNNs in the real world \cite{Lipton2016, highstake3}. Furthermore, DL remains vulnerable to adversarial perturbations, the imperceptible changes crafted to disrupt DL models' decisions-making process \cite{adversarial, reviewadver1, reviewadver2}. 

Therefore, to utilize DL's capability in high-stakes domains, we must look into DNNs’ decision making process. As DNNs successively transform inputs into final predictions, probing hidden layers' responses can shed light on their operations. Earlier studies showed that shallow layers (i.e, layers close to the input layer) encode simple low-level features, whereas deep layers (i.e., layers close to the output layer) encode high-level features \cite{olah2017feature, grad3, carter2019activation}. The high level features sometimes include human interpretable features, and the low-level features in the shallow layers are similar to stimuli, to which biological neurons in the visual systems of the early stage would respond \cite{Olshausen}. 

Then, two important questions arise. How do individual layers evolve? And how do they contribute to DNNs' decisions? Multiple studies have been dedicated to answer these questions, which has led to a subfield of DL named `explainable AI’, and fall into three categories. The first line of studies estimated inputs’ influence on DNNs' decisions. Specifically, several studies utilized the gradient-based analysis \cite{GradCAM, IntegratedGradient, Gradient2}, the layerwise relevance propagation that assigns individual neurons' relevance using backpropagation-like rules \cite{LRP} and the occlusion analysis that measured input patches’ influence \cite{grad3}. The second line of studies focused on analyzing hidden layer representations. The feature visualization \cite{feature, carter2019activation} inverted DNNs to identify optimal features that can drive certain hidden layers. Testing with concept activation vectors (TCAV) used examples with and without human-annotated concepts to train a secondary model to learn concept vectors \cite{TCAV}. Once the secondary model is trained, it can evaluate the influence of concept vectors on DL models. Finally, the third line of studies evaluated direct relationships between hidden layer representations and DNNs' decisions \cite{alain2018understanding, library}. 

To better understand the functional codes underlying DNNs' operations, we used unsupervised learning to analyze hidden layer representations. Our study is based on the assumption that the number of functional codes is much smaller than the dimension of input feature space, and the mapping between inputs and functional codes should be many-to-one. If each input can be mapped onto a unique functional code, DNNs can effectively become memory cells that do not generalize their predictions on unknown inputs (queries). We also assumed that the functional codes would appear repeatedly during DNNs' operations. With these assumptions in mind, we used the self-organizing map (SOM) \cite{som2, Kohonen:2007}, an unsupervised learning algorithm, to probe hidden layer representations. Our analyses suggest that shallow layers in DNNs map inputs onto extremely homogeneous codes, and deep layers successively map these homogeneous codes in shallow layers onto more diverse codes. This unexpected finding reminded us of the fact that the convex and concave lenses in telescopes collect light and enhance image resolution.  

Then, why do DL models first map inputs onto homogeneous codes? We propose that the homogeneous codes (or condensed feature space, in which all codes are closely located) can enable DL models to efficiently generalize from training examples. Although this telescope-like operation can enhance the generalization ability, we found evidence suggesting that it can also underlie DL's vulnerabilities to adversarial perturbations.

\begin{figure}
  \centering
  \includegraphics[width=0.8\linewidth]{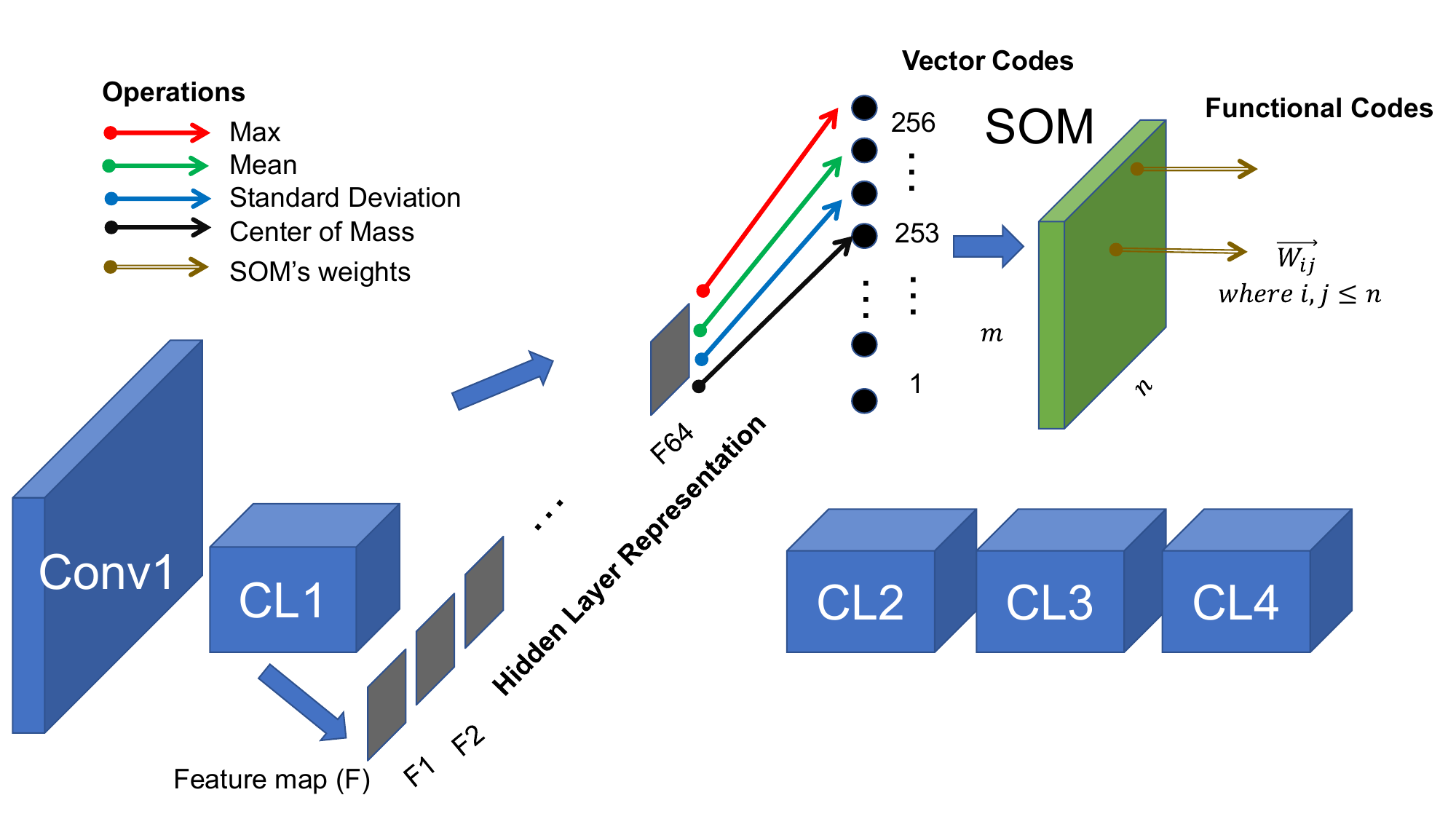}
  \caption{Schematics of SOM-based analysis. We display ResNet18 as an example. The same approach is applied to ResNet50, VGG19 and DenseNet121. We convert individual feature maps’ responses to vector codes using max, mean, standard deviation and center of mass. Vector codes are abstract representations of feature map outputs (hidden layer representations) and are inputs to a SOM, which is used to identify the functional codes.}
  \label{som_diagram}
\end{figure}

\section{Approach}
In principle, DNNs are trained to generate many-to-one mappings. For instance, classifiers are trained to map high dimensional inputs into a small number of classes. Based on the fact that DL models contain identical functional blocks in series, we hypothesize that internal modules of DL models may also produce many-to-one like mappings. More specifically, we hypothesize that 1) a group of inputs is mapped onto the same internal code in a hidden layer, and 2) the memberships of functional groups mapped onto the same internal codes depend on layers. For instance, inputs $A$ and $B$ can be mapped onto the same code in layer $n$, but inputs $A$ and $B$ are not necessarily mapped onto the same code in layer $m$. 

This hypothesis has an important implication for explainable DL. If we decode such internal many-to-one like mappings, we may use them to decompose DL models’ decisions into functional components.  Then, how do we confirm or reject this hypothesis? If many-to-one like mappings are internally created and used for DL models’ functions, the internal codes in hidden layers will be grouped together. Thus, we need to check if the hidden layer codes/representations are grouped together, and if not, we can reject this hypothesis. In our study, we use the self-organizing map (SOM), an unsupervised learning algorithm, to test the hidden layer codes. 

We refer to the raw hidden layer outputs as `hidden layer representations', which are known to be highly variable and sensitive to input pixel intensities. We refer to the internal representations as `functional codes', which mediate semantic meanings of inputs. To discover functional codes used by DL models, we first convert hidden layer representations into more abstract vector codes by characterizing feature maps' outputs with their statistical properties. 

\subsection{Vector Codes of feature maps}
Convolution filters produce 2-dimensional outputs (feature maps), which reflect the spatial distribution of inputs. In modern DNNs, dimensions of all the features produced by these filters are often too big. If we take activations of all feature maps by flattening them into 1D vectors, the final vectors reflecting individual layers contain a great amount of stochastic noise, making the analysis overly sensitive to statistical fluctuation or biases that may be irrelevant to semantic information of inputs. Earlier studies employed various dimension reduction algorithms to analyze features; see \cite{rathore2021topoact, purvine2022experimental, alain2018understanding} for example. Specifically, Alain and Bengio \cite{alain2018understanding} used two dimension reduction methods. First, responses were randomly sampled from a subset of spatial locations of feature maps. This `random sampling’ was also used by other studies \cite{rathore2021topoact, purvine2022experimental}. Second, 2D pooling was used to reduce the number of features to the number of channels. We note that random sampling or naive 2D pooling may result in some crucial information of inputs to be lost, especially in shallow layers, in which feature maps have large spatial dimensions (both height and width). To minimize the loss of information, we coded the individual feature maps by taking max, mean, standard deviation and center of mass of activations (Fig. \ref{som_diagram}), which can help us partially preserve spatial information of the feature maps and make our analysis insensitive to some stochastic/spurious noises independent of inputs' semantic meanings. These abstract codes will be referred to as vector codes below.

\subsection{Identification of functional codes}
We assume that internal representations can be decomposed into functional codes and noise, and noises can be averaged out by native operations (e.g., summations/convolutions) occurring inside DNNs. As SOM \cite{som2, Kohonen:2007} can identify representative inputs and their relationships in high dimensional feature space, we used it to identify effective functional codes from vector codes. Fig. \ref{som_diagram} illustrates our experiment procedure. We first recorded hidden layer representations to obtain vector codes and then trained a SOM on vector codes to identify functional codes. 

In this study, we analyzed 1) four convolutional (Conv) networks, ResNet18, ResNet50 \cite{resnet}, DenseNet121 \cite{huang2018densely} and VGG19 \cite{vgg19}, trained on ImageNet \cite{imagenet} and 2) Vision Transformers \cite{dosovitskiy2021imageworth16x16words, steiner2022trainvitdataaugmentation} trained on ImageNet, 3) ResNet50 trained on CIFAR10, CIFAR100 \cite{cifar10100} and SVHN\cite{Netzer2011ReadingDI}. The four Conv consist of composite blocks. ResNets consist of 4 composite blocks (\cite{resnet}), which will be referred to as the composite layer (CL) below (Fig. \ref{som_diagram}). Similarly, DenseNet121 and VGG19 also have functional blocks naturally corresponding to ResNet's CLs. As these functional blocks can represent early, middle and late processing stages, we analyzed the functional codes in them using Timm, an open source pytorch library \cite{rw2019timm} and its feature extractor API. Specifically, we obtained hidden layer representations from examples randomly chosen from ImageNet \cite{imagenet}. Then, the hidden layer representations were converted to 1-dimensional vector codes using max, mean, standard deviation and center of mass. Following this rule, the dimension of the vector codes are 4 times bigger than the number of channels. For instance, we obtained 256, 512, 1024, 2048-dimensional vector codes from ResNet18.  

ViT has different architecture from Conv layers. ViT uses multi-head attention layer instead of convolutional layers, and it consists of 12 embedding layers, which has the same dimension (a.k.a, embedding dissension). Notably, as ViT’s classification token captures compressed information on the entire image, we analyzed classification tokens in all 12 embedding layers. That is, we did not employ vector codes when it comes to ViT. 

\subsection{SOM and its training to determine functional codes} \label{som_training}
SOM consists of units distributed over 2-dimensional grids \cite{som2, Kohonen:2007}. Each unit stores a single weight vector, whose length is the same as that of an input. During training, each input is compared to the weights stored in the units, and the best matching unit (BMU) is identified. Then, the weight of BMU and its neighbors are updated according to the learning rule (Eq. \ref{som1}). 
\begin{equation} \label{som1}
    W_v(s+1)=W_v(s)+ exp\left( -\frac{D^2}{2\sigma(s)^2+\epsilon} \right)\times\alpha(s)\times (x-W_v(s)) 
\end{equation}
, where $s$ denotes training index; $u$ and $v$ denote the best matching unit (BMU) and a unit to be updated, respectively; where $W$ denotes a weight; where $\alpha$ is a learning rate; $x$ is an input; where $D=||x-W_u||$, distance metric; $\epsilon$ is added for numerical stability.

With this update rule, SOM’s weights move towards inputs, but the degree of changes depend on the locations of the units. BMU or its neighboring units update weights strongly, but the units far from BMU update them less strongly. Notably, SOM has two properties. First, the weights of SOM tend to be similar to the inputs and represent exemplary inputs after training. Second, the neighboring units store similar weights, and consequently the distance between BMUs reflects the similarity between the inputs. That is, after training, the similarity between the two inputs can be measured with the distance between their corresponding BMUs. We used these properties to analyze the functional codes.

We used two sets of 50,000 inputs randomly drawn from ImageNet \cite{imagenet}. The first set (SOM-tr) was built with the ImageNet train set, and the second set (SOM-val), with the ImageNet validation set. Both sets contain 50 examples per each ImageNet class. Only SOM-tr was used to train SOMs, and SOM-val was used to probe their responses to unseen examples. For individual inputs from the two sets, we obtained vector codes from 4 CLs of ResNets and grouped them according to layers (i.e., CL). The vector codes from the same layer were used to train and test a SOM. As there are 4 CLs, we built and trained 4 independent SOMs. A wide range of distance metrics have been used for SOMs in the earlier studies, and we selected the cosine distance (CD, Eq. \ref{cd}) to measure the similarity between SOMs’ weights and a vector code. SOMs were constructed with the periodic boundary to avoid boundary effects observed during training. With the periodic boundary condition, SOM effectively became a 2-dimensional torus, and the distance was measured accordingly. 
\begin{equation} \label{cd}
CD(u, v)=1-\frac{u \cdot v}{\| u \| \|v \|}
\end{equation}
We used an open source python library QuickSOM \cite{quicksom} to construct SOM with a square grid. In the experiment, we tested SOMs of 4 different sizes. They contained 10-by-10, 20-by-20, 30-by-30 or 40-by-40 nodes at square grids. As we did not observe significant dependency on SOMs’ sizes, the analyses below are based on SOMs consisting of 20-by-20 units, unless stated otherwise. Table \ref{table:som} shows the hyperparameters selected for the experiments, and the terminologies are adopted from QuickSom. 

\begin{table}[htb]
\caption{We list the parameters selected for SOM. $m$ and $n$ denote the height and width, respectively. In the experiment, we used a square grid ($m=n$). $\sigma$ and $\alpha$ denote their initial values, and they decay over time. We used `exp' decay mode supported by Quicksom \cite{quicksom}}\label{table:som}

\label{table1}
\centering
\begin{tabular}{|c|c|}
\hline
Parameter      & value \\ \hline
$m$ (height) & 10,20,30,40 \\
$n$ (width)  & 10,20,30,40 \\
$\sigma$   & 10     \\
$\alpha$    & 0.1  \\
decay      & exp   \\ \hline
\end{tabular}
\end{table}

\subsection{Analysis of functional codes }\label{functional code_analysis}
After training, we considered weights of trained SOMs as effective functional codes of DL models (Fig. \ref{som_diagram}) and used them to address two questions. First, can SOMs learn functional codes? In other words, can the weights of SOMs mediate semantic meanings of inputs (in our analysis, the labels of inputs)? Second, what properties do functional codes in 4 CLs have? Regarding the first question, we obtained top-10 BMUs corresponding to individual vector codes elicited by images in SOM-tr and SOM-val and grouped them according to inputs’ labels. If SOMs can learn the functional codes, BMUs corresponding to the vector codes elicited by the same class inputs would be close to one another. For the second question, we studied BMUs’ relationships to one another by measuring the Euclidean distance between BMUs and CD (Eq. \ref{cd}) between their weights. 

\begin{figure}
  \centering
  \includegraphics[width=1\linewidth]{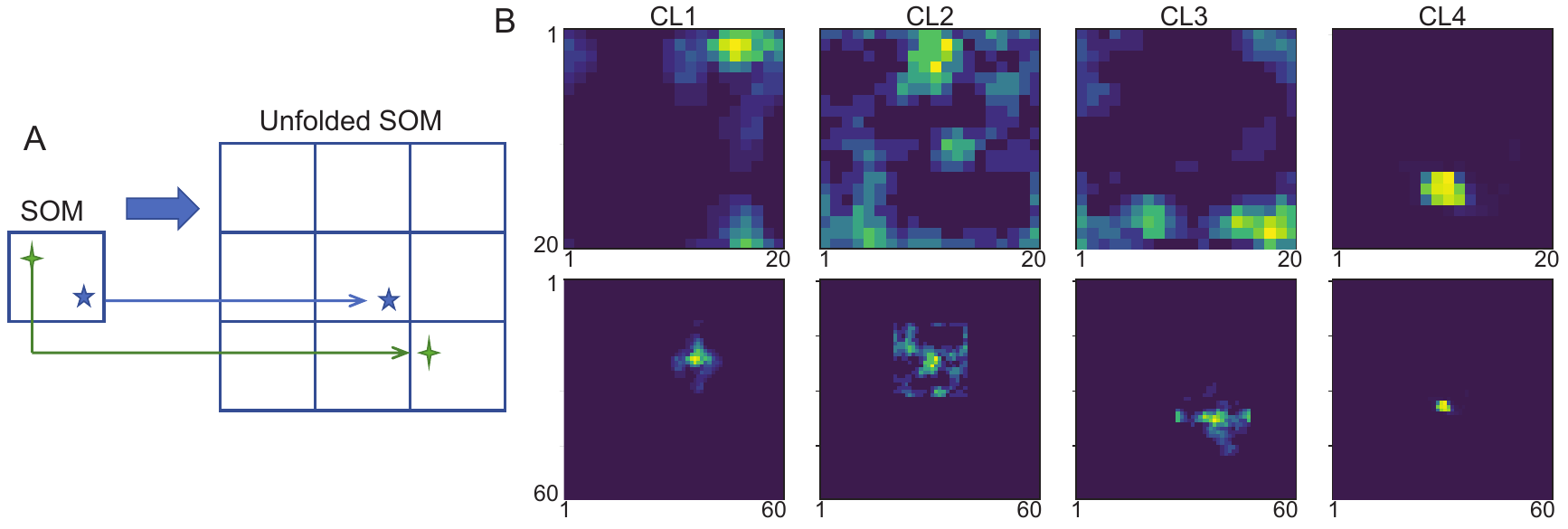}
  \caption{Unfolding SOM. When Euclidean distances between two SOM units or the class centers are estimated, all BMUs are transferred to the unfolded SOM. (A), Torus distance. To estimate the torus distance, all BMUs are mapped onto 9 identical grids. (B), Examples of BMUs in the original (top row) and unfolded (bottom row) SOMs. The columns show BMUs from CL1, CL2, CL3 and CL4, respectively. The class center is the center of BMUs’ mass in the unfolded SOMs shown in the bottom row. It should be noted that SOMs in the bottom row are three times larger than the original SOMs in the top row.}
  \label{unfold}
\end{figure}

As a constructed SOM is a torus (due to the periodic boundary condition), we measured the shortest distance between SOMs' nodes in a torus. Specifically, the trained SOM was mapped onto the center of 9 grids. When we estimated Euclidean distance between the two SOM units, one of the two units was mapped onto the center grid, and another, onto one  of 9 grids (Fig. \ref{unfold}A), in which the shortest distance is available. In our analysis, we also identified the class centers using SOM’s BMUs. In doing so, for each class, we first determined which unit was frequently chosen as BMU, while the same class inputs from SOM-tr were fed to ImageNet models. Then, all BMUs of the same class examples were mapped onto one of 9 grids (Fig. \ref{unfold}A), in which the distance to the most frequently chosen BMU was the shortest. The class center is the center of BMUs’ mass in this unfolded SOM (see the bottom row of Fig. \ref{unfold}B for examples).  

\section{Results}
Using SOM, we analyze hidden representations of Conv networks trained on ImageNet, CIFAR10, CIFAR100 and SVHN (Sec.\ref{res1}) and Vision transformer, ViT,  (Sec. \ref{res2}).  Importantly, our analyses raise the possibility that both Conv networks and ViT maps inputs onto homogeneous codes in earlier hidden layers. We discuss the implications of these surprising findings for the generalization ability and the vulnerabilities to adversarial perturbations (Sec. \ref{res3} and \ref{res4}).

\subsection{Functional codes of Conv networks}\label{res1}
\begin{figure}
  \centering
  \includegraphics[width=1\linewidth]{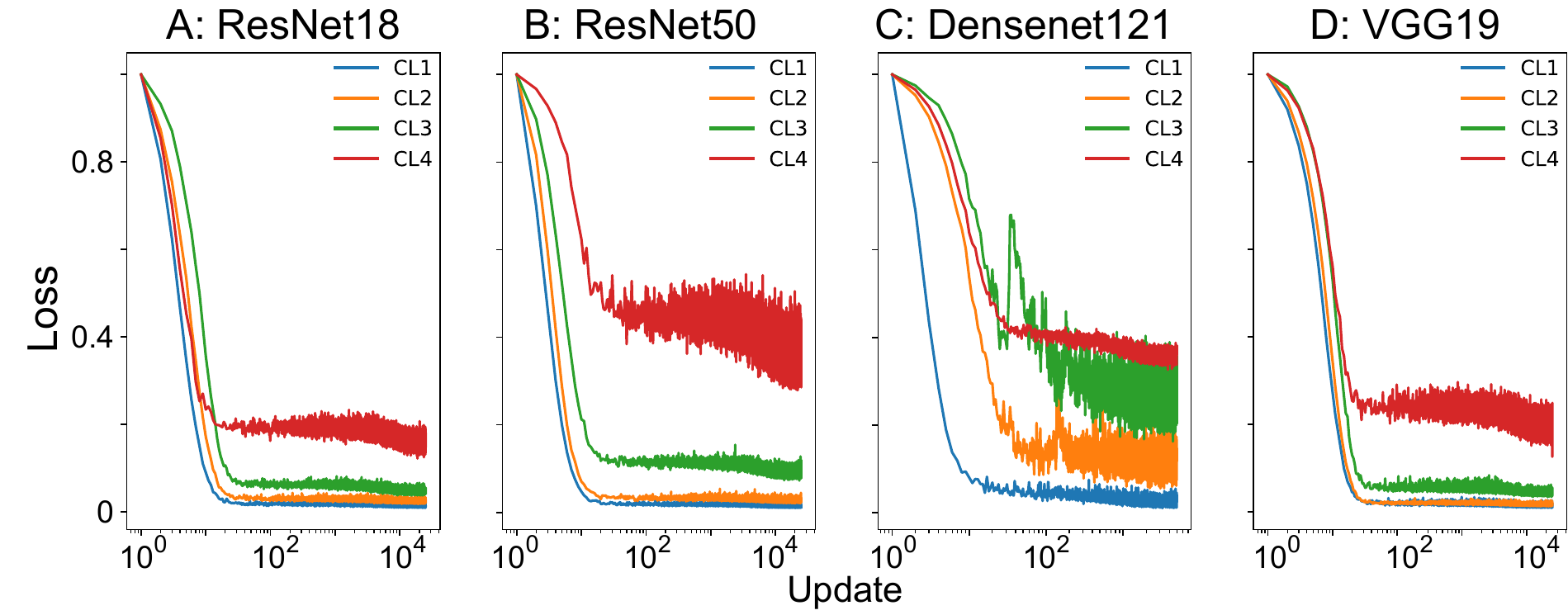}

  \caption{Loss function of SOMs trained on 4 models' vector codes. Blue, orange, green and red line denote CL1, CL2, CL3 and CL4, respectively. (A)-(D), Time courses of SOM training of ResNet18, ResNet50, DenseNet121 and VGG19. }
  \label{loss}
\end{figure}

We first obtain vector codes (Sec.\ref{functional code_analysis}) from CL1, 2, 3 and 4 and train SOMs on them for 5 epochs. It should be noted that 4 SOMs are independently trained on CL1, 2, 3 and 4 to obtain layer-specific functional codes. During training, we use 10 as a batch size for ResNet18, 50 and VGG19 and 50 for DenseNet121.  As shown in Fig. \ref{loss}A, the loss functions of SOMs, trained on ResNet18’s vector codes, decline initially and then become saturated at different levels depending on layers. We also observe the equivalent results with ResNet50, DenseNet121 and VGG19 (Fig. \ref{loss}B-D). For each model, loss is higher in CL4 than earlier layers. As SOMs quantize inputs, these results suggest that hidden layer responses are diverse in CL4 but quite homogeneous in CL1. 

\begin{figure*}\label{density}
  \centering
  \includegraphics[width=1\textwidth]{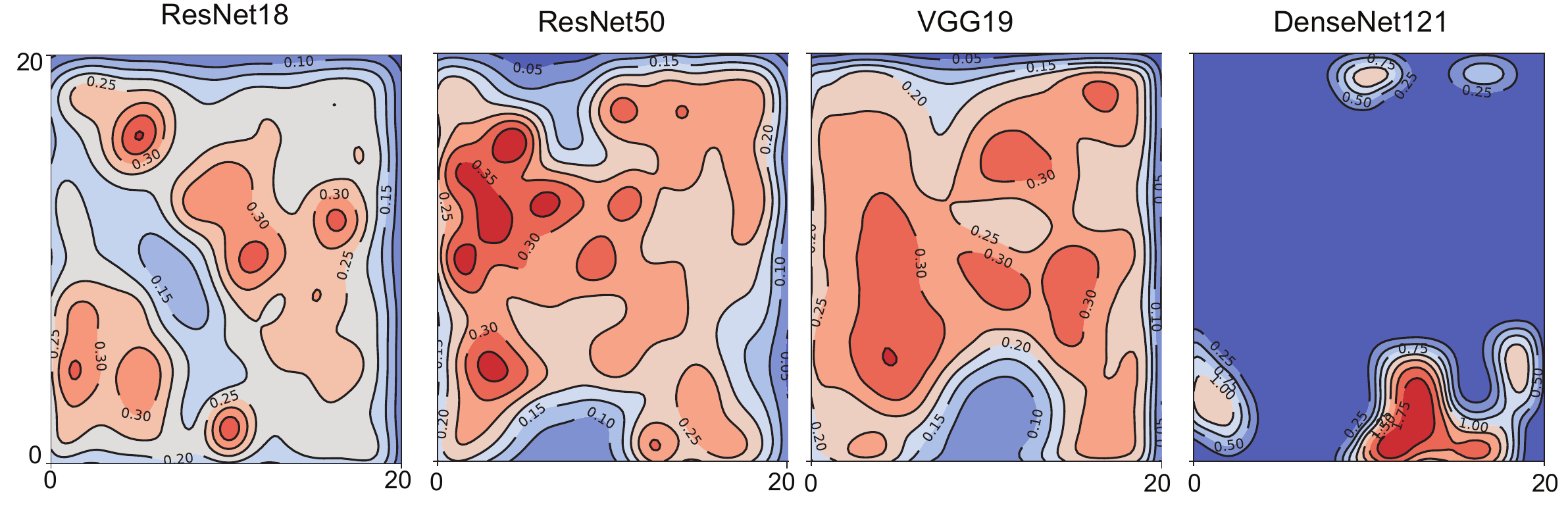}

  \caption{Density of BMUs. 4 panels show BMUs of ResNet18, ResNet50, VGG19 and DenseNet121, respectively.  We estimate the densities of BMUs using Gaussian kernels \cite{2020SciPy-NMeth} (see also Supplementary Fig. 1). For the sake of brevity, we show BMUs of CL2 only. For all 4 layers, see Supplementary Fig. 1. We note that the density per area is very low. For illustration purposes only, we scale up the density by 100 times.}
  \label{gauss}
\end{figure*}

We evaluate the density of BMUs for all SOM-tr in its 2D grid. As shown in Fig. \ref{gauss}, BMUs are spread inhomogeneously, and some units are rarely selected as BMUs. These results suggest that hidden layer responses are similar to one another, and they can be mapped onto a finite number of representative response patterns, which can be the functional codes in DNNs. We further ask if these codes are correlated with the labels (semantic meaning) of inputs.

\begin{figure*}
  \centering
  \includegraphics[width=0.8\linewidth]{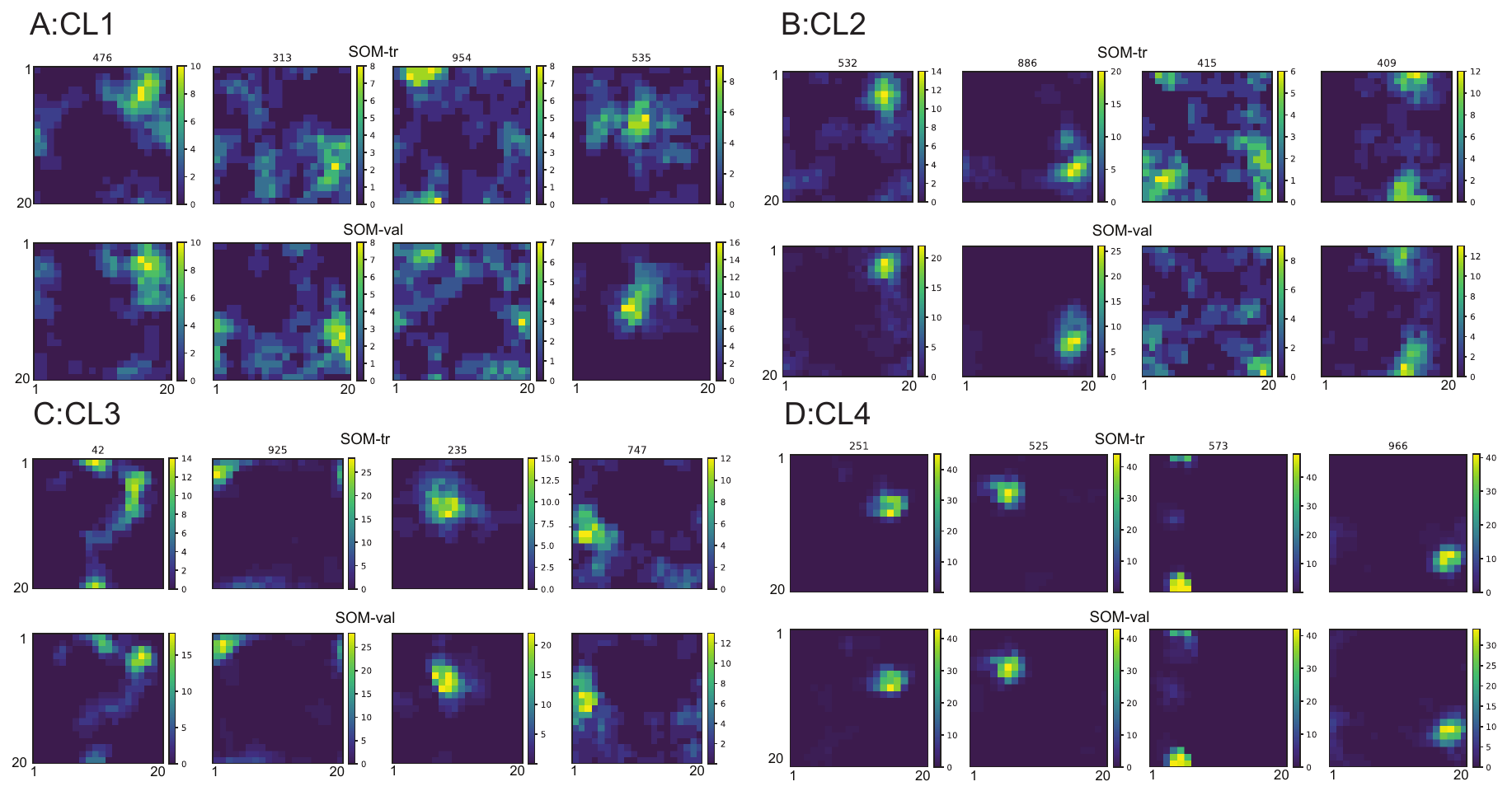}
\caption{BMUs of vector codes of ResNet18. For each input, we collect vector codes from CL1-CL4 of ResNet18 and estimate their 10 BMUs in the trained SOM. (A), BMUs in CL1. (B), BMUs in CL2, (C), BMUs in CL3. (D), BMUs in CL4. The top panels show BMUs of examples in SOM-tr, which are used to train SOMs. The bottom panels show BMUs of examples in SOM-val, which are not used to train SOMs. Each column shows BMUs of inputs in the same class; the number above the panels denotes their class IDs in the ImageNet.}
  \label{TrainVal_ResNet18}
\end{figure*}

Specifically, we test if the hidden layer responses elicited by the same class inputs could be mapped onto the neighboring BMUs. As shown in Fig. \ref{TrainVal_ResNet18}, BMUs of the same class inputs are clustered together. Notably, the spatial spreads of BMUs are relatively bigger in shallow layers but smaller in deep layers. Furthermore, we observe that SOM-val, which is not used to train SOMs, elicits a similar set of units (Fig. \ref{TrainVal_ResNet18}); we find equivalent results from SOM trained on ResNet50 (Supplementary Fig. 2). These results suggest that SOMs can learn semantically meaningful codes, and their correlations to the labels (i.e., classes) of inputs become stronger in deep layers. 

\begin{figure*}
  \centering
  \includegraphics[width=0.9\linewidth]{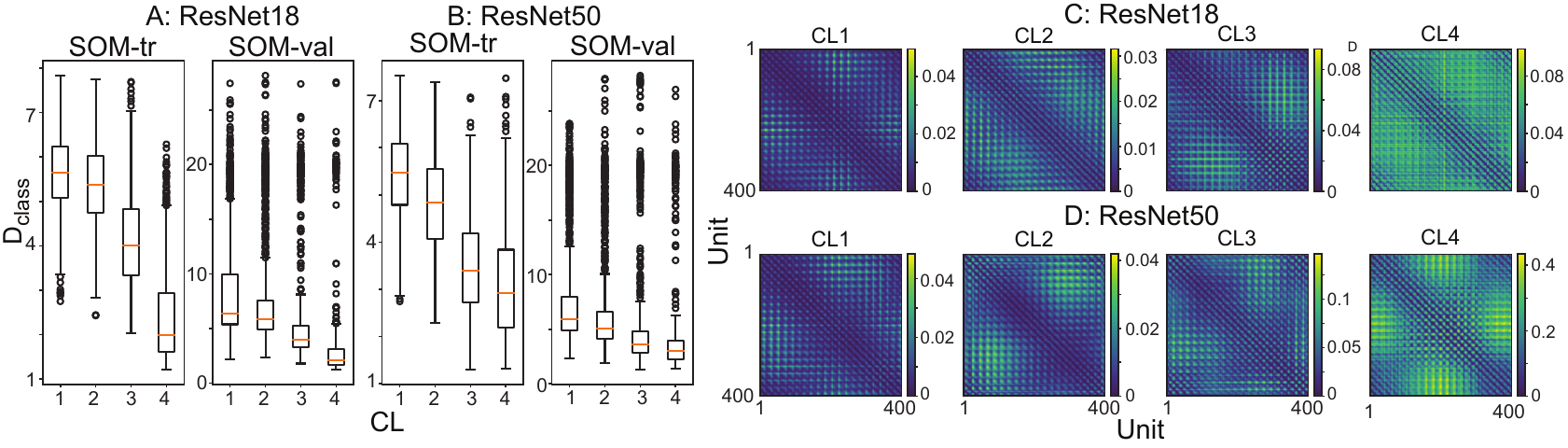}
  \caption{$D_{class}$, Euclidean distance between individual BMUs and their class centers. For each ImageNet class, we first obtain the class center and calculate the mean distance between BMUs and the class center. To show the statistical properties of all 1000 classes, we use the box plot. (A), $D_{class}$ measured in ResNet18 using SOM-tr (left) and SOM-val (right). The open circles denote the outliers (i.e., classes significantly different from others). We observe the outliers more frequently in SOM-val, which is not used to train SOMs. (B), $D_{class}$ in ResNet50. (C), Cosine distance (Eq. \ref{cd}) between the weights of SOMs trained on vector codes from 4 CLs of ResNet18. (D), Cosine distance between weights of SOMs trained on ResNet50. We display all pairs of 400 units in the 20-by-20 SOM.}
  \label{BoxPlot}
\end{figure*}

To quantify this observation, we evaluate the Euclidean distance ($D_{class}$) between individual BMUs of the same class inputs and their centers, as the distance between BMUs reflect their similarity. In doing so, we first identify the class center (i.e., the centroids of BMUs of the same class inputs) and calculate the $D_{class}$ in the 2D torus (see section \ref{functional code_analysis} and Fig. \ref{unfold}). In this analysis, we calculate the average distances on  class-by-class basis. Fig. \ref{BoxPlot}A and B show the distributions of distances of all 1000 classes. As shown in the figure, the distances to the class centers are much shorter in deep layers than those in shallow layers. The distances measured by unseen examples (from ImageNet validation set) indicate the same trend. We also ask if $D_{class}$s are correlated with ResNet's answers and find that $D_{class}$ is bigger when ResNet makes an incorrect prediction than when ResNet makes a correct prediction; see supplementary Text and Supplementary Fig. 3. 

\begin{figure}
  \centering
  \includegraphics[width=0.8\linewidth]{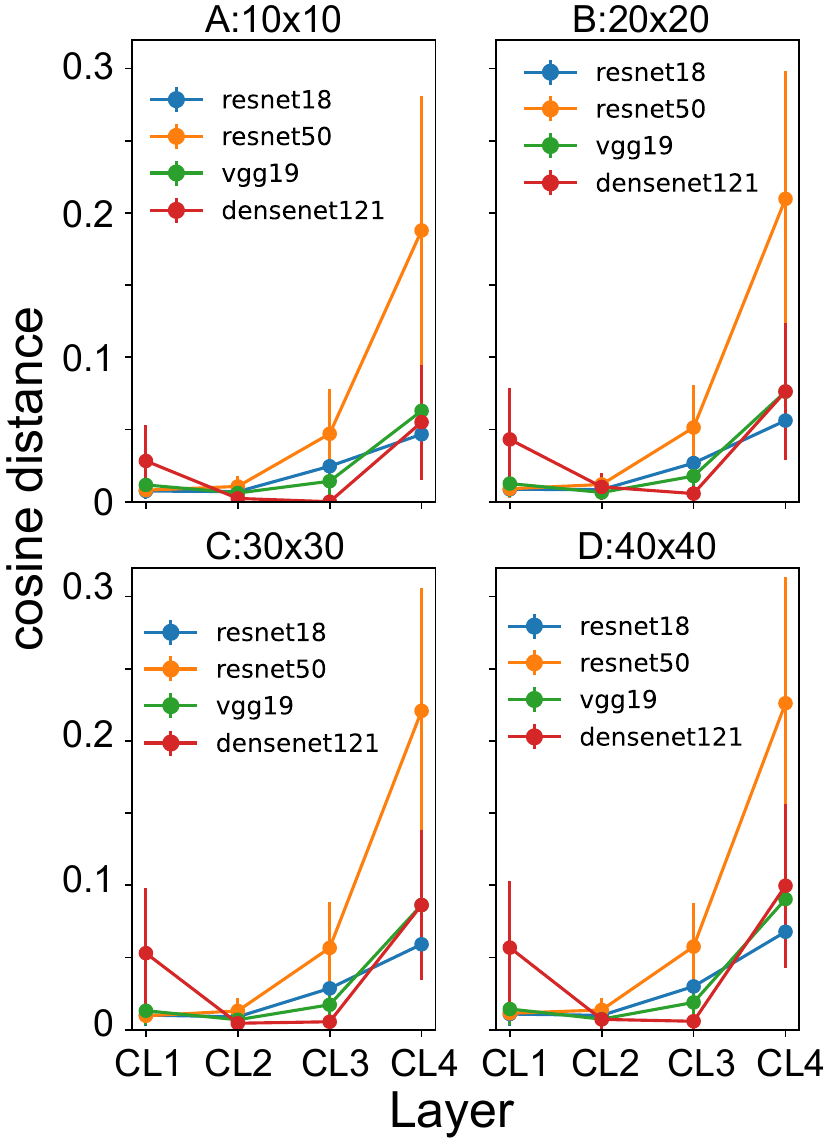}
\caption{Cosine distance between class centers' weights in all 4 layers. We measure cosine distances from SOMs with $10 \times 10$, $20 \times 20$, $30 \times 30$ and $40 \times 40$ nodes, and they are shown in (A)-(D), respectively. In each panel, the color denotes the model analyzed.}
  \label{class_distance}
\end{figure}

As the results above suggest that SOMs can learn semantically meaningful functional codes and store them as weights, we further analyze the weights of SOMs. First, we measure the similarity among trained weights of SOMs by using cosine distance. We observe that the weights of SOMs trained on hidden layer responses in CL1 and CL2 are remarkably similar and that their similarity decreases in CL3 and CL4 (Fig. \ref{BoxPlot} C and D).  Second, we evaluate the similarity between class centers. We first identify class centers in the 2D grid of SOMs and determine the nearest neighbor units. We randomly choose 100 class centers (more precisely, center units) and compute the cosine distance between their weights. We display them for all 4 Conv models in (Fig. \ref{class_distance}). 

\begin{figure}[htb]
  \centering
  \includegraphics[width=1\linewidth]{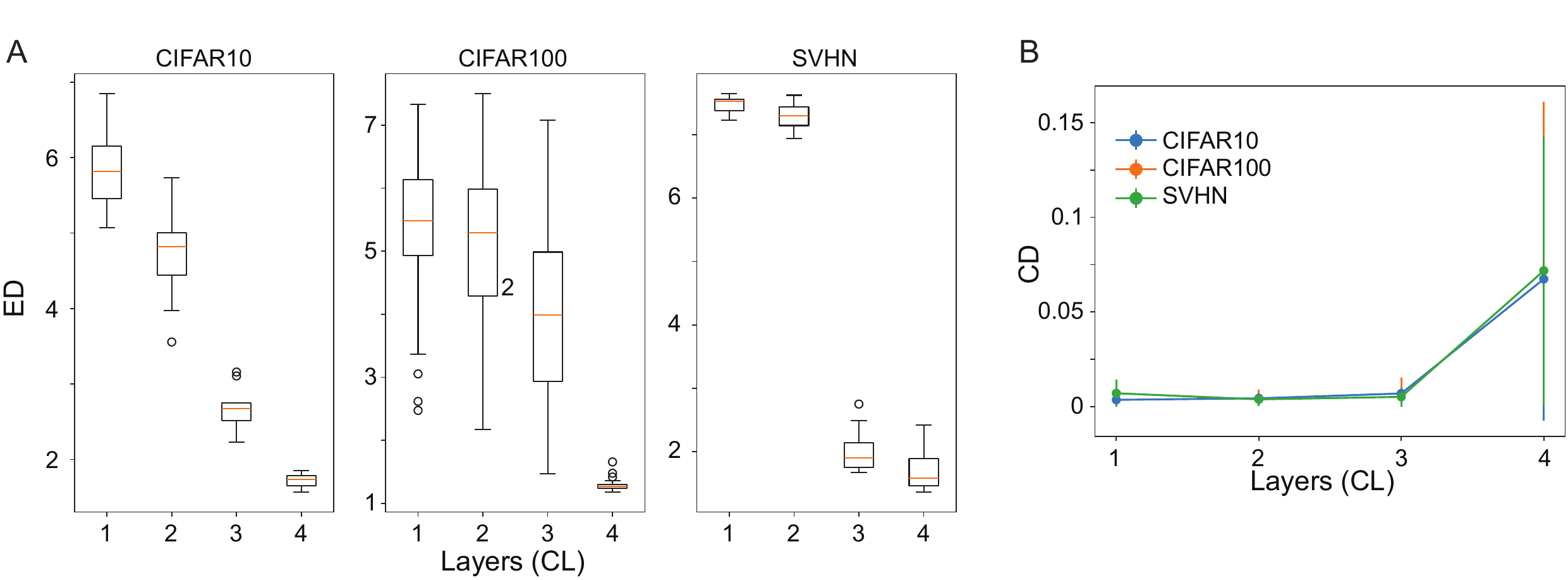}
\caption{Analysis of ResNet50 trained on CIFAR10, CIFAR100 and SVHN. (A), Euclidean distance between datapoints and class centers depending on the dataset. (B), Cosine distance of class centers depending on the dataset.}
\label{other}
\end{figure}

These results raise the possibility that Conv networks trained on ImageNet project inputs to condensed feature space, in which datapoints are similar to one another. We test if this trend is also valid for other datasets by training SOMs on hidden layer responses of ResNet50 trained on CIFAR10, 100 \cite{krizhevsky2009learning} and SVHN \cite{netzer2011reading} and find the equivalent results (Fig. \ref{other}). 

\subsection{Functional codes in Vision Transformer} \label{res2}
\begin{figure}[htb]
  \centering
  \includegraphics[width=1\linewidth]{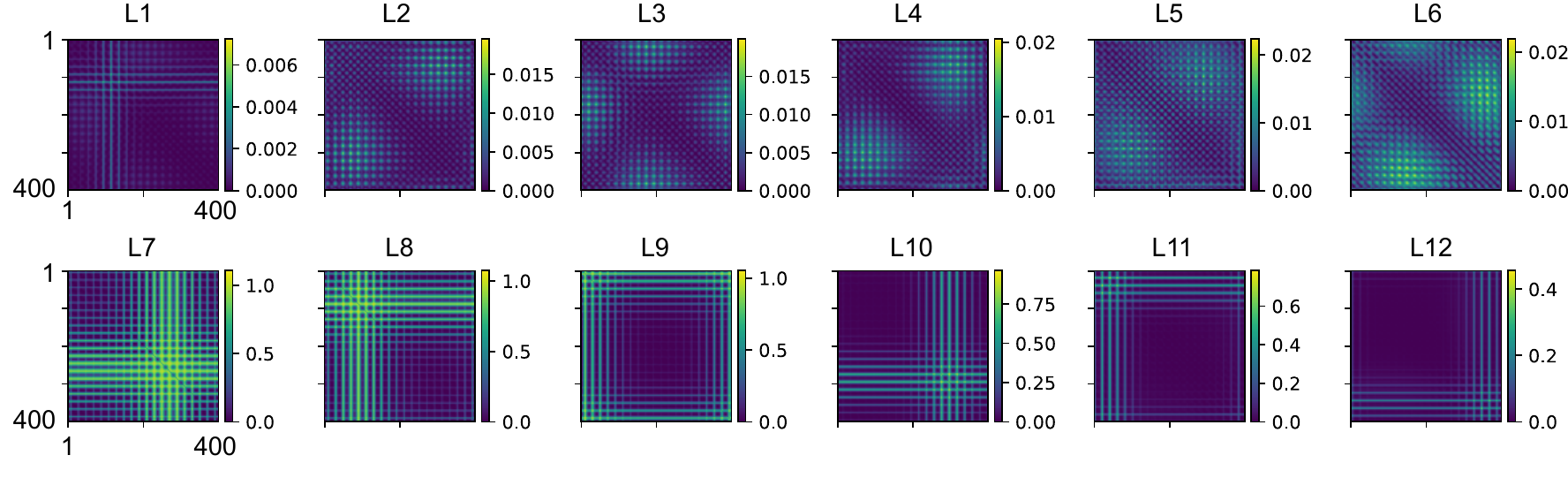}
\caption{Cosine distance between weights of SOM trained on ViT's embedding layer.}
\label{vit_weight}
\end{figure}

\begin{figure}[htb]
  \centering
  \includegraphics[width=1\linewidth]{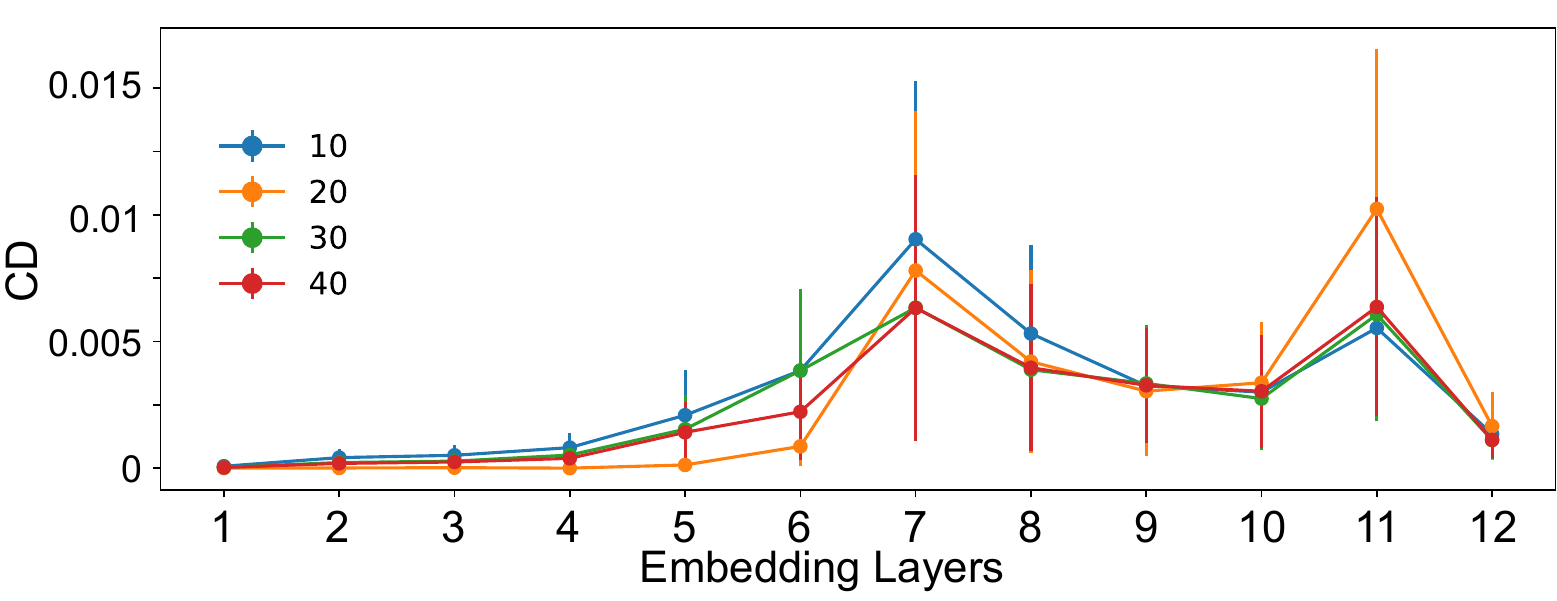}
\caption{Cosine distance between class centers observed from 12 ViT embedding layers. The color denotes the size of SOM. For instance, ‘10’ denotes a SOM consisting of $10 \times 10$ nodes.}
\label{vit_cd} 
\end{figure}

Next, we ask if Vision Transformer (ViT) share the same property. ViT inserts additional classification `CLS' token in each encoding layer, which can code an entire visual scene (input), and the final CLS token is used for classification. Thus, we train SOMs on the CLS tokens in all 12 encoding layers of ViT. The cosine distances between weights of trained SOMs are shown in Fig. \ref{vit_weight}, and the cosine distances between 100 class centers are shown in Fig. \ref{vit_cd}. These results are consistent with those observed in CNNs, suggesting that ViT also maps inputs to condensed feature space. 

\subsection{Sampling efficient generalization} \label{res3}
The results above can lead to a question: why do DL models utilize compressed feature space, in which all inputs are mapped onto datapoints closely located to one another? The exact reason may remain unclear for now, but we argue that compressed feature space allows DL to estimate the landscape of error (LoE) with a small number of samples. In principle, a DL model $f$ reduces prediction error $\epsilon$ for each pair of an input $x$ and a label $y$ using a gradient descent. Thus, alternatively, we may consider a DL model as a function, $f(x, y)=\epsilon$, which predicts error $\epsilon$, given the input $x$ and the label $y$. During inference, it chooses the $y$ that can minimize $\epsilon$. With this in mind, we note that a prediction on test examples $x$ can be inferred via Taylor expansion Eq.\ref{eq3}, if there is a sufficient number of training examples $x_t$ in the vicinity of $x$. 
\begin{equation}\label{eq3}
f(x, y)=\Sigma_{t=1,n}f(x_t, y_t)+\frac{df(x_t,y_t)}{dx}(x-x_t).     
\end{equation}
, where $n$ is the number of neighboring training examples $x_t$. For all labels $y$, it can predict an error $\epsilon$ to given $x$. As visual images are not homogeneous, the approximation may not hold for the raw images. However, our analysis raises the possibility that DL models first project (in-distribution) inputs into compressed feature space, in which Eq. \ref{eq3} will hold quite well, as multiple training examples $x_t$ exist in the vicinity of test examples $x$.  That is, compressed feature space suggested by our analyses can make DL models generalize answers efficiently based on training examples in a sampling efficient way.

\subsection{Generalization ability and vulnerabilities of adversarial perturbations: the two sides of the same coin} \label{res4}
The compressed feature space may allow DL models to generalize training examples effectively, but it may also make DL models vulnerable to small perturbations. Since all inputs are mapped onto neighboring datapoints in feature space of early layers, a small perturbation on datapoints in the compressed feature space may trigger a move from one answer to another, which indicates DL models' vulnerabilities to adversarial perturbations.  To address this hypothesis, we turn to adversarial perturbations known to change DL models’ predictions with minimal perturbations on inputs. Specifically, we evaluate the influence of adversarial inputs on functional codes in ResNet18 and ResNet50 by using SOMs. As the distances between BMUs reflect the similarity between inputs, we measure the Euclidean distance between clean BMUs and adversarial BMUs. When adversarial perturbations change the functional codes significantly, the distances will be farther. In this analysis, we also measure the influence of random perturbations as a control condition. 

\begin{figure}[htb]
  \centering
  \includegraphics[width=1.0\linewidth]{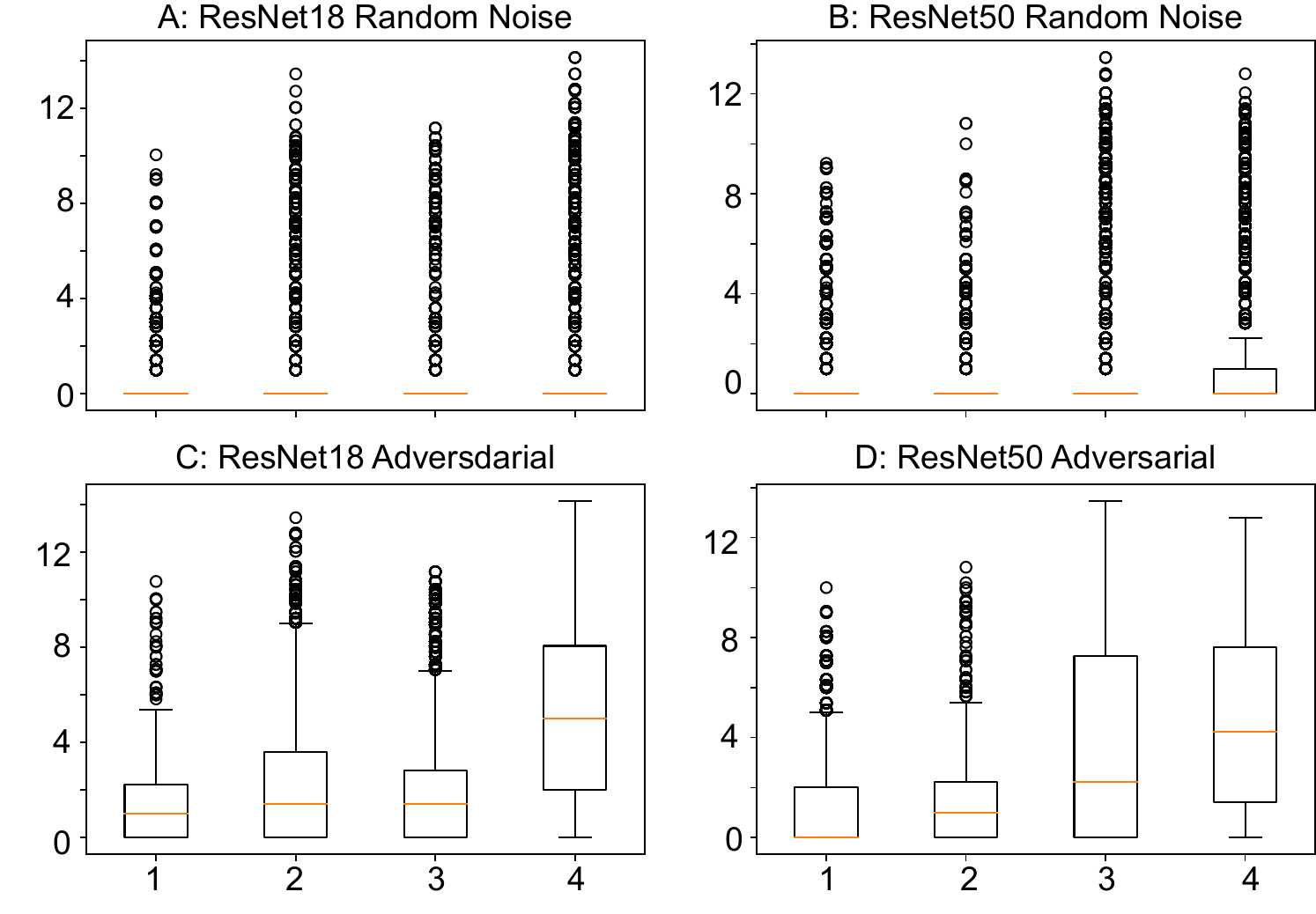}
\caption{Influence of adversarial and random perturbations on functional codes in ResNets. We measure the Euclidean distance between clean BMUs and perturbed BMUs. For each pair of clean and perturbed inputs, we record 10 BMUs from both images and calculate the Euclidean distance. (A), Euclidean distance between clean and random noise BMUs when ResNet18 is analyzed. (B), the same as (A) but the model is ResNet50. (C), Euclidean distance between clean and adversarial inputs when ResNet18 is analyzed. (D), the same as (C) but the model is ResNet50.}
  \label{class_distance_adv}
\end{figure}

We randomly choose 500 images from ImageNet and perturb them using projected gradient attack (PGD) \cite{pgd} and an open-source adversarial attack library \cite{ding2019advertorch}. We restrain perturbation strength $\epsilon$ to be smaller than 4/255; see Table \ref{tables1} for other hyperparameters used in the experiment. Random noises, used as a control condition here, are added to the same 500 images and are independently drawn from a uniform distribution between 0 and 4/255. Although the magnitudes of adversarial and random perturbations are comparable, their influences on ResNet18 and 50 are strikingly different. The accuracy of ResNet18 and ResNet50 prediction drop from 77.8\% and 93.0 \% to 3.6\% and 6.6\%, respectively, whereas random noises do not change the models’ prediction accuracy (77.8\% and 93.0 \%). 

\begin{figure}[htb]
  \centering
  \includegraphics[width=1.0\linewidth]{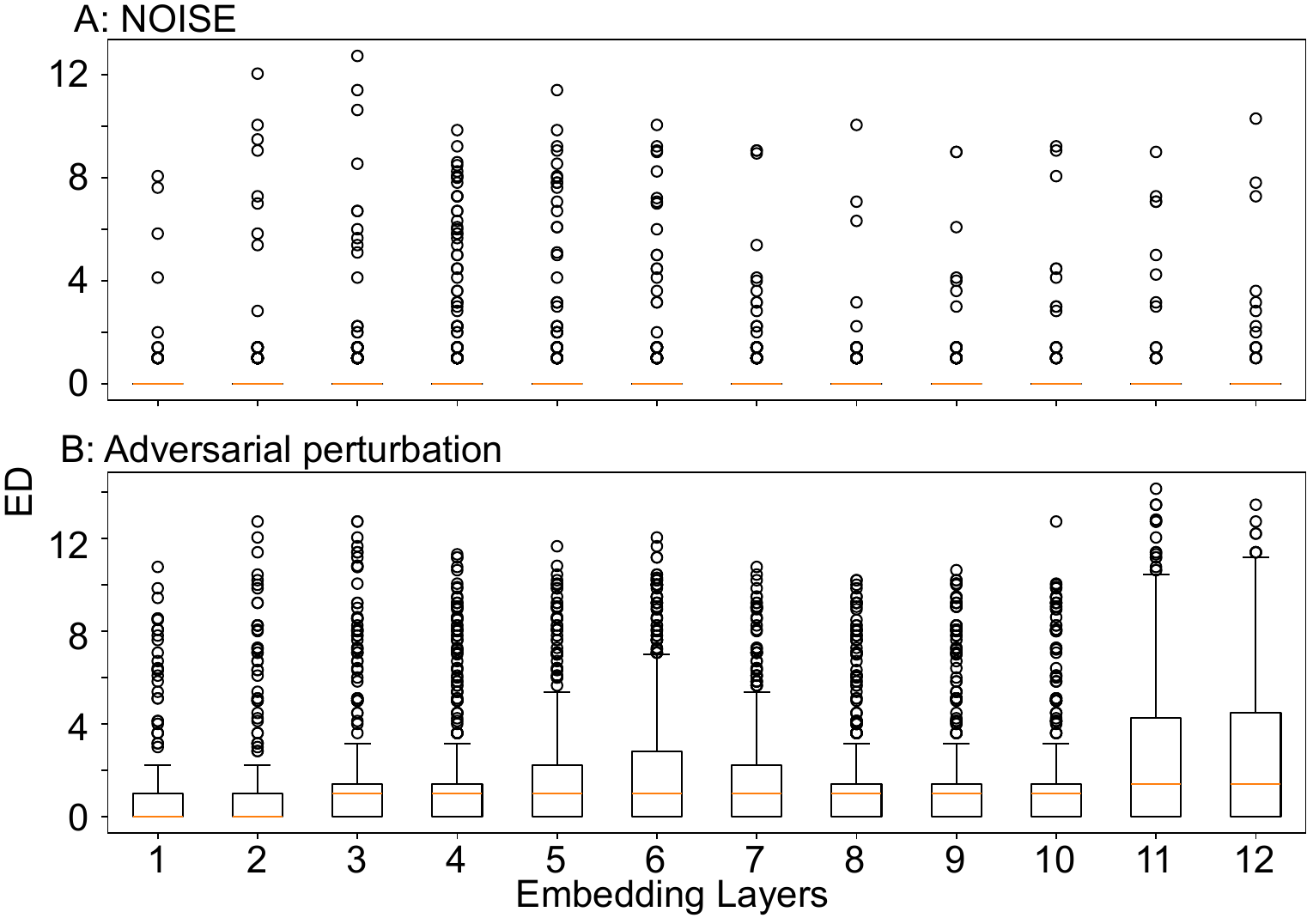}
\caption{Influence of adversarial and random perturbations on functional codes in ViT. We measure the Euclidean distance between clean BMUs and perturbed BMUs. For each pair of a clean and a perturbed input, we record 10 BMUs from both images and calculate the Euclidean distance. }
  \label{vit_adv}
\end{figure}

\begin{table}[htb]
\caption{Parameters used to craft adversarial images.}
\label{tables1}
\centering
\begin{tabular}{|l|l|}
\hline
Arguments (name in advertorch) & Value                               \\ \hline
Loss function (loss\_fn)       & CrossEntropyLoss   \\
Maximum perturbation (eps, $\epsilon$)     & $\frac{4}{255}$ \\
Iteration (nb\_iter)           & 50                          \\
Update step size (eps\_iter)   & $\frac{4}{255} \times \frac{2}{100} $                           \\
rand\_int                      & True                                \\
targeted                       & False                                \\
clip\_min                      & 0.0                                 \\
clip\_max                      & 1.0                                 \\
norm                           & $L_{inf}$                     \\
\hline
\end{tabular}
\end{table}
For each pair of a clean image and a perturbed image, we obtain 10 best BMUs from both clean and perturbed images. Figs. \ref{class_distance_adv} A and B show the distances between 10 pairs of BMUs elicited by clean and random noise inputs. As shown in the figures, random noise inputs do not have substantial influence on all 4 layers. By contrast, adversarial inputs do change the functional codes in all 4 layers, but in significantly different degrees. In CL1, the distances observed are very small, but they are noticeably bigger in CL4. We repeat the same analysis with ViT and observed the equivalent results (Fig. \ref{vit_adv}).

\section{Discussion}
Based on our results, we propose that DNNs' operations closely resemble telescopes’ operations and that the compressed feature space, in which all features/codes are homogeneous, may be a double-edged sword which readily enhances generalization ability but creates DNNs’ vulnerabilities to adversarial perturbations.
\subsection{Links to earlier works}
Producing generalizations is one of DNNs’ key capabilities, and an earlier study suggested that their vulnerabilities to adversarial perturbations are connected to the generalization of non-essential features \cite{NEURIPS2019_e2c420d9}. Our analyses are consistent with their finding and further suggest that such erroneous generalization can easily occur due to compressed feature space. The model stitching \cite{bansal2021revisiting} suggested that shallow layers from more powerful DNNs can boost less powerful DNNs’ performance. Our functional telescope hypothesis suggests that shallow layers’ essential function is to create condensed feature space and thus they may be exchangeable between models, which leads to the possibility of model stitching. 
\subsection{Functional Telescope hypothesis}
Inspired by our two observations that functional codes in shallow layers are homogeneous (Figs. \ref{BoxPlot}, \ref{class_distance}), and that adversarial attacks use small perturbations in shallow layers to induce strong perturbations in functional codes in deeper layers (Figs.\ref{class_distance_adv} and \ref{vit_adv}), we propose our functional telescope hypothesis that DL uses shallow layers to map high-dimensional inputs into extremely homogeneous codes and use deeper layers to gradually convert them into decision variables central for accurate predictions on the class of inputs. Supplementary  Fig. 4 summarizes our hypothesis, and we argue that this hypothesis can explain that small perturbations producing significant influences in deeper layers. 

\bibliography{main}

\begin{thebibliography}{10}

\bibitem{Lecun2015}
Yann Lecun, Yoshua Bengio, and Geoffrey Hinton.
\newblock {Deep learning}.
\newblock {\em Nature}, 521(7553):436--444, 2015.

\bibitem{deep_review1}
Ajay Shrestha and Ausif Mahmood.
\newblock Review of deep learning algorithms and architectures.
\newblock {\em IEEE Access}, 7:53040--53065, 2019.

\bibitem{deep_review2}
Saptarshi Sengupta, Sanchita Basak, Pallabi Saikia, Sayak Paul, Vasilios
  Tsalavoutis, Frederick Atiah, Vadlamani Ravi, and Alan Peters.
\newblock A review of deep learning with special emphasis on architectures,
  applications and recent trends.
\newblock {\em Knowledge-Based Systems}, 194:105596, 2020.

\bibitem{Lipton2016}
Zachary~C. Lipton.
\newblock {The Mythos of Model Interpretability}.
\newblock In {\em ICML WHI}, 2016.

\bibitem{highstake3}
Cynthia Rudin.
\newblock {Please Stop Explaining Black Box Models for High Stakes Decisions}.
\newblock In {\em NIPS Workshop}, 2018.

\bibitem{adversarial}
Christian Szegedy, Wojciech Zaremba, Ilya Sutskever, Joan Bruna, Dumitru Erhan,
  Ian Goodfellow, and Rob Fergus.
\newblock Intriguing properties of neural networks, 2013.

\bibitem{reviewadver1}
Anirban Chakraborty, Manaar Alam, Vishal Dey, Anupam Chattopadhyay, and Debdeep
  Mukhopadhyay.
\newblock Adversarial attacks and defences: A survey, 2018.

\bibitem{reviewadver2}
Xiaoyong Yuan, Pan He, Qile Zhu, and Xiaolin Li.
\newblock Adversarial examples: Attacks and defenses for deep learning.
\newblock {\em IEEE Transactions on Neural Networks and Learning Systems},
  30(9):2805--2824, 2019.

\bibitem{olah2017feature}
Chris Olah, Alexander Mordvintsev, and Ludwig Schubert.
\newblock Feature visualization.
\newblock {\em Distill}, 2017.
\newblock https://distill.pub/2017/feature-visualization.

\bibitem{grad3}
Matthew~D. Zeiler and Rob Fergus.
\newblock Visualizing and understanding convolutional networks.
\newblock {\em CoRR}, abs/1311.2901, 2013.

\bibitem{carter2019activation}
Shan Carter, Zan Armstrong, Ludwig Schubert, Ian Johnson, and Chris Olah.
\newblock Activation atlas.
\newblock {\em Distill}, 2019.
\newblock https://distill.pub/2019/activation-atlas.

\bibitem{Olshausen}
Bruno Olshausen and David Field.
\newblock Emergence of simple-cell receptive field properties by learning a
  sparse code for natural images.
\newblock {\em Nature}, 381:607--9, 07 1996.

\bibitem{GradCAM}
Ramprasaath~R. Selvaraju, Michael Cogswell, Abhishek Das, Ramakrishna Vedantam,
  Devi Parikh, and Dhruv Batra.
\newblock Grad-{CAM}: Visual explanations from deep networks via gradient-based
  localization.
\newblock {\em International Journal of Computer Vision}, 128(2):336--359, oct
  2019.

\bibitem{IntegratedGradient}
Mukund Sundararajan, Ankur Taly, and Qiqi Yan.
\newblock Axiomatic attribution for deep networks.
\newblock In Doina Precup and Yee~Whye Teh, editors, {\em Proceedings of the
  34th International Conference on Machine Learning}, volume~70 of {\em
  Proceedings of Machine Learning Research}, pages 3319--3328. PMLR, 06--11 Aug
  2017.

\bibitem{Gradient2}
Marco Ancona, Enea Ceolini, Cengiz Öztireli, and Markus Gross.
\newblock Towards better understanding of gradient-based attribution methods
  for deep neural networks.
\newblock In {\em International Conference on Learning Representations}, 2018.

\bibitem{LRP}
Gr{\'e}goire Montavon, Alexander Binder, Sebastian Lapuschkin, Wojciech Samek,
  and Klaus-Robert M{\"u}ller.
\newblock {\em Layer-Wise Relevance Propagation: An Overview}, pages 193--209.
\newblock Springer International Publishing, Cham, 2019.

\bibitem{feature}
Chris Olah, Alexander Mordvintsev, and Ludwig Schubert.
\newblock Feature visualization.
\newblock {\em Distill}, 2017.
\newblock https://distill.pub/2017/feature-visualization.

\bibitem{TCAV}
Been Kim, Martin Wattenberg, Justin Gilmer, Carrie Cai, James Wexler, Fernanda
  Viegas, and Rory sayres.
\newblock Interpretability beyond feature attribution: Quantitative testing
  with concept activation vectors ({TCAV}).
\newblock In Jennifer Dy and Andreas Krause, editors, {\em Proceedings of the
  35th International Conference on Machine Learning}, volume~80 of {\em
  Proceedings of Machine Learning Research}, pages 2668--2677. PMLR, 10--15 Jul
  2018.

\bibitem{alain2018understanding}
Guillaume Alain and Yoshua Bengio.
\newblock Understanding intermediate layers using linear classifier probes,
  2018.

\bibitem{library}
Jung~Hoon Lee.
\newblock Library network, a possible path to explainable neural networks,
  2019.

\bibitem{som2}
T.~Kohonen.
\newblock The self-organizing map.
\newblock {\em Proceedings of the IEEE}, 78(9):1464--1480, 1990.

\bibitem{Kohonen:2007}
T.~Kohonen and T.~Honkela.
\newblock {K}ohonen network.
\newblock {\em Scholarpedia}, 2(1):1568, 2007.
\newblock revision \#127841.

\bibitem{rathore2021topoact}
Archit Rathore, Nithin Chalapathi, Sourabh Palande, and Bei Wang.
\newblock Topoact: Visually exploring the shape of activations in deep
  learning, 2021.

\bibitem{purvine2022experimental}
Emilie Purvine, Davis Brown, Brett Jefferson, Cliff Joslyn, Brenda Praggastis,
  Archit Rathore, Madelyn Shapiro, Bei Wang, and Youjia Zhou.
\newblock Experimental observations of the topology of convolutional neural
  network activations, 2022.

\bibitem{resnet}
Kaiming He, Xiangyu Zhang, Shaoqing Ren, and Jian Sun.
\newblock Deep residual learning for image recognition, 2015.

\bibitem{huang2018densely}
Gao Huang, Zhuang Liu, Laurens van~der Maaten, and Kilian~Q. Weinberger.
\newblock Densely connected convolutional networks, 2018.

\bibitem{vgg19}
Karen Simonyan and Andrew Zisserman.
\newblock Very deep convolutional networks for large-scale image recognition.
\newblock In {\em International Conference on Learning Representations}, 2015.

\bibitem{imagenet}
Jia Deng, Wei Dong, Richard Socher, Li-Jia Li, Kai Li, and Li~Fei-Fei.
\newblock Imagenet: A large-scale hierarchical image database.
\newblock In {\em 2009 IEEE conference on computer vision and pattern
  recognition}, pages 248--255. Ieee, 2009.

\bibitem{dosovitskiy2021imageworth16x16words}
Alexey Dosovitskiy, Lucas Beyer, Alexander Kolesnikov, Dirk Weissenborn,
  Xiaohua Zhai, Thomas Unterthiner, Mostafa Dehghani, Matthias Minderer, Georg
  Heigold, Sylvain Gelly, Jakob Uszkoreit, and Neil Houlsby.
\newblock An image is worth 16x16 words: Transformers for image recognition at
  scale, 2021.

\bibitem{steiner2022trainvitdataaugmentation}
Andreas Steiner, Alexander Kolesnikov, Xiaohua Zhai, Ross Wightman, Jakob
  Uszkoreit, and Lucas Beyer.
\newblock How to train your vit? data, augmentation, and regularization in
  vision transformers, 2022.

\bibitem{cifar10100}
Alex Krizhevsky.
\newblock Learning multiple layers of features from tiny images.
\newblock Technical report, 2009.

\bibitem{Netzer2011ReadingDI}
Yuval Netzer, Tao Wang, Adam Coates, A.~Bissacco, Bo~Wu, and A.~Ng.
\newblock Reading digits in natural images with unsupervised feature learning.
\newblock 2011.

\bibitem{rw2019timm}
Ross Wightman.
\newblock Pytorch image models.
\newblock \url{https://github.com/rwightman/pytorch-image-models}, 2019.

\bibitem{quicksom}
Vincent Mallet, Michael Nilges, and Guillaume Bouvier.
\newblock {quicksom: Self-Organizing Maps on GPUs for clustering of molecular
  dynamics trajectories}.
\newblock {\em Bioinformatics}, 37(14):2064--2065, 11 2020.

\bibitem{2020SciPy-NMeth}
Pauli Virtanen, Ralf Gommers, Travis~E. Oliphant, Matt Haberland, Tyler Reddy,
  David Cournapeau, Evgeni Burovski, Pearu Peterson, Warren Weckesser, Jonathan
  Bright, St{\'e}fan~J. {van der Walt}, Matthew Brett, Joshua Wilson, K.~Jarrod
  Millman, Nikolay Mayorov, Andrew R.~J. Nelson, Eric Jones, Robert Kern, Eric
  Larson, C~J Carey, {\.I}lhan Polat, Yu~Feng, Eric~W. Moore, Jake
  {VanderPlas}, Denis Laxalde, Josef Perktold, Robert Cimrman, Ian Henriksen,
  E.~A. Quintero, Charles~R. Harris, Anne~M. Archibald, Ant{\^o}nio~H. Ribeiro,
  Fabian Pedregosa, Paul {van Mulbregt}, and {SciPy 1.0 Contributors}.
\newblock {{SciPy} 1.0: Fundamental Algorithms for Scientific Computing in
  Python}.
\newblock {\em Nature Methods}, 17:261--272, 2020.

\bibitem{krizhevsky2009learning}
Alex Krizhevsky, Geoffrey Hinton, et~al.
\newblock Learning multiple layers of features from tiny images.
\newblock 2009.

\bibitem{netzer2011reading}
Yuval Netzer, Tao Wang, Adam Coates, Alessandro Bissacco, Bo~Wu, and Andrew~Y
  Ng.
\newblock Reading digits in natural images with unsupervised feature learning.
\newblock 2011.

\bibitem{pgd}
Aleksander Madry, Aleksandar Makelov, Ludwig Schmidt, Dimitris Tsipras, and
  Adrian Vladu.
\newblock Towards deep learning models resistant to adversarial attacks, 2017.

\bibitem{ding2019advertorch}
Gavin~Weiguang Ding, Luyu Wang, and Xiaomeng Jin.
\newblock {AdverTorch} v0.1: An adversarial robustness toolbox based on
  pytorch.
\newblock {\em arXiv preprint arXiv:1902.07623}, 2019.

\bibitem{NEURIPS2019_e2c420d9}
Andrew Ilyas, Shibani Santurkar, Dimitris Tsipras, Logan Engstrom, Brandon
  Tran, and Aleksander Madry.
\newblock Adversarial examples are not bugs, they are features.
\newblock In H.~Wallach, H.~Larochelle, A.~Beygelzimer, F.~d\textquotesingle
  Alch\'{e}-Buc, E.~Fox, and R.~Garnett, editors, {\em Advances in Neural
  Information Processing Systems}, volume~32. Curran Associates, Inc., 2019.

\bibitem{bansal2021revisiting}
Yamini Bansal, Preetum Nakkiran, and Boaz Barak.
\newblock Revisiting model stitching to compare neural representations.
\newblock In A.~Beygelzimer, Y.~Dauphin, P.~Liang, and J.~Wortman Vaughan,
  editors, {\em Advances in Neural Information Processing Systems}, 2021.

\end{thebibliography}

\bibliographystyle{unsrt}

\clearpage
\setcounter{page}{1}

\setcounter{table}{0}
\renewcommand{\tablename}{Supplementary Table}

\setcounter{figure}{0}
\renewcommand{\figurename}{Supplementary Figure}

\setcounter{page}{1}

\setcounter{table}{0}
\renewcommand{\tablename}{Supplementary Table}

\setcounter{figure}{0}
\renewcommand{\figurename}{Supplementary Figure}
\subsection*{Supplemental Text}
We test if ResNets utilize functional codes for decision-making by estimating the correlations between functional codes and ResNets’ decisions. In this test, we group functional codes depending on whether ResNets make correct predictions of the inputs’ labels. If functional codes are linked to ResNets’ decisions, BMUs storing functional codes would be close to the centers of the class predicted by ResNets. If this assumption holds, when ResNets make correct predictions, BMUs storing functional codes would be close to the centers of the labels. By contrast, when ResNets make incorrect predictions, they would be close to the centers of the classes different from the labels. Based on this line of thought, we expect the distance (Euclidean distance, ED) between BMUs and the label centers to be bigger when ResNets’ make incorrect predictions than when they make incorrect predictions. In the analysis, we estimate the centers of the labels using SOM-tr and test ED between BMUs and the label centers using SOM-val; this is consistent with other analyses, in which the class centers are estimated using SOM-tr only. As shown in Supplemental Fig. 3, ED is bigger on average when ResNets make mistakes.

\subsection*{Supplemental Figures}

\begin{figure*}[h]
  \centering
  \includegraphics[width=0.9\textwidth]{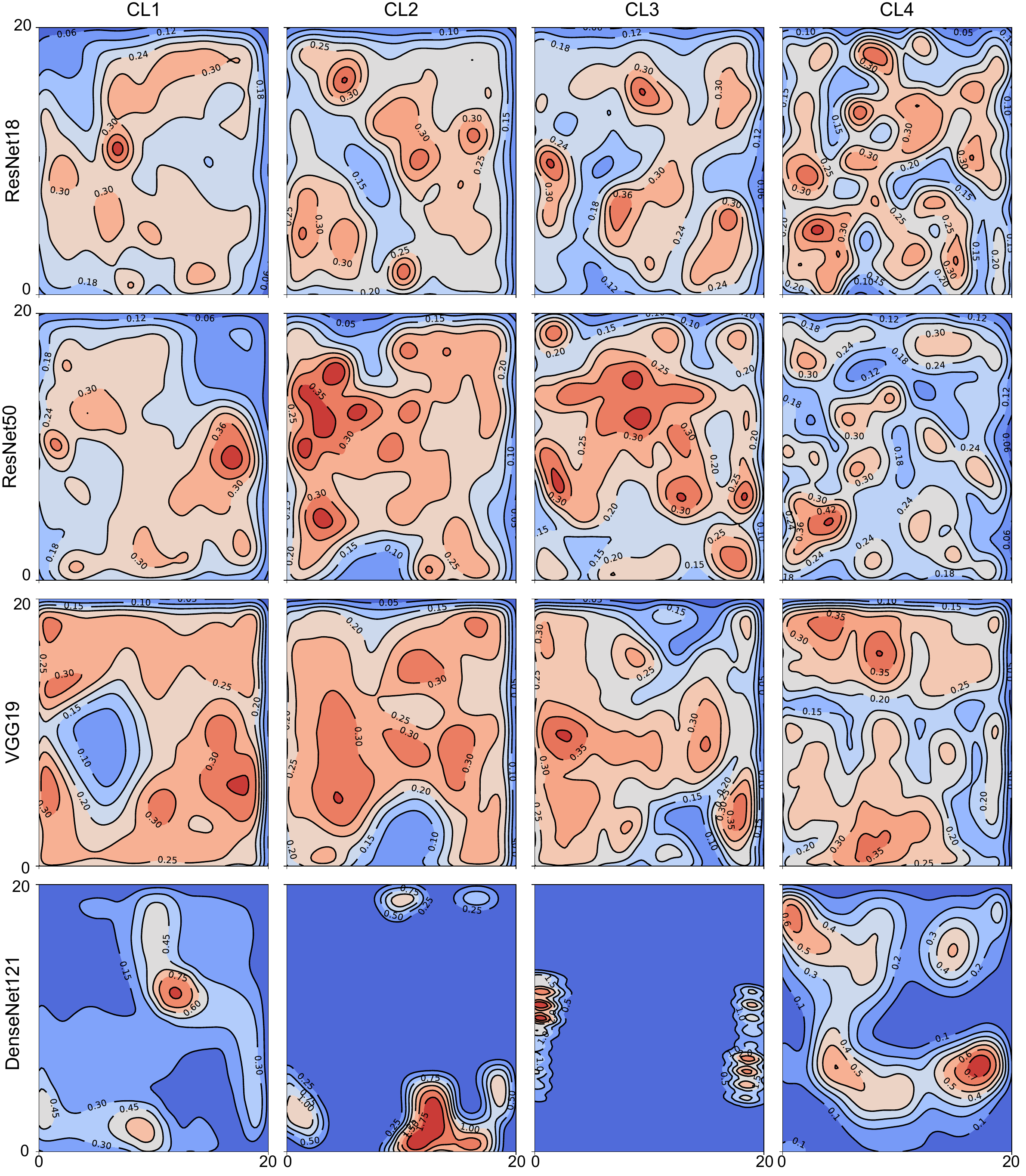}

  \caption{Density of BMUs. Four rows show BMUs of ResNet18, ResNet50, VGG19 and DenseNet121, respectively. Each column shows BMUs from one of 4 CLs (see the name of CL shown above the column). We note that the density per area is very low. For illustration purpose only, we scaled the density by 100 times.}
  \label{gauss_all}
\end{figure*}

\begin{figure*}[h]
  \centering
  \includegraphics[width=0.9\linewidth]{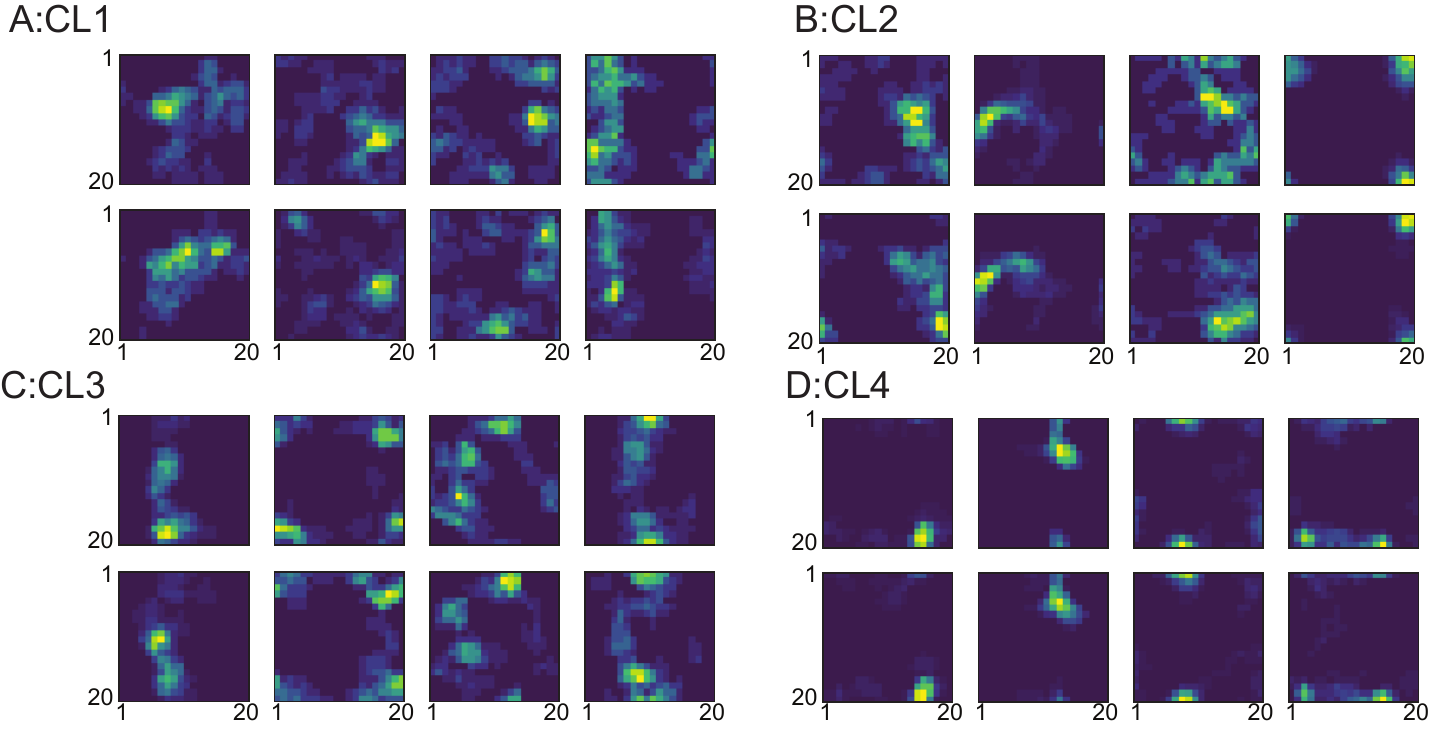}
\caption{BMUs of vector codes obtained from ResNet50. The panels show BMUs of SOMs trained on ResNet50. (A), BMUs of vector codes in CL1. The top row shows the BMUs of SOM-tr, whereas the bottom row shows the BMUs of SOM-val. Each column shows BMUs of the inputs in the same class. (B), the same as (A), but vector codes were obtained from CL2. (C), the same as (A), but vector codes were obtained from CL3.(D), the same as (A), but vector codes were obtained from CL4.}
  \label{TrainVal_ResNet50}
\end{figure*}

\begin{figure*}
  \centering
  \includegraphics[width=0.6\linewidth]{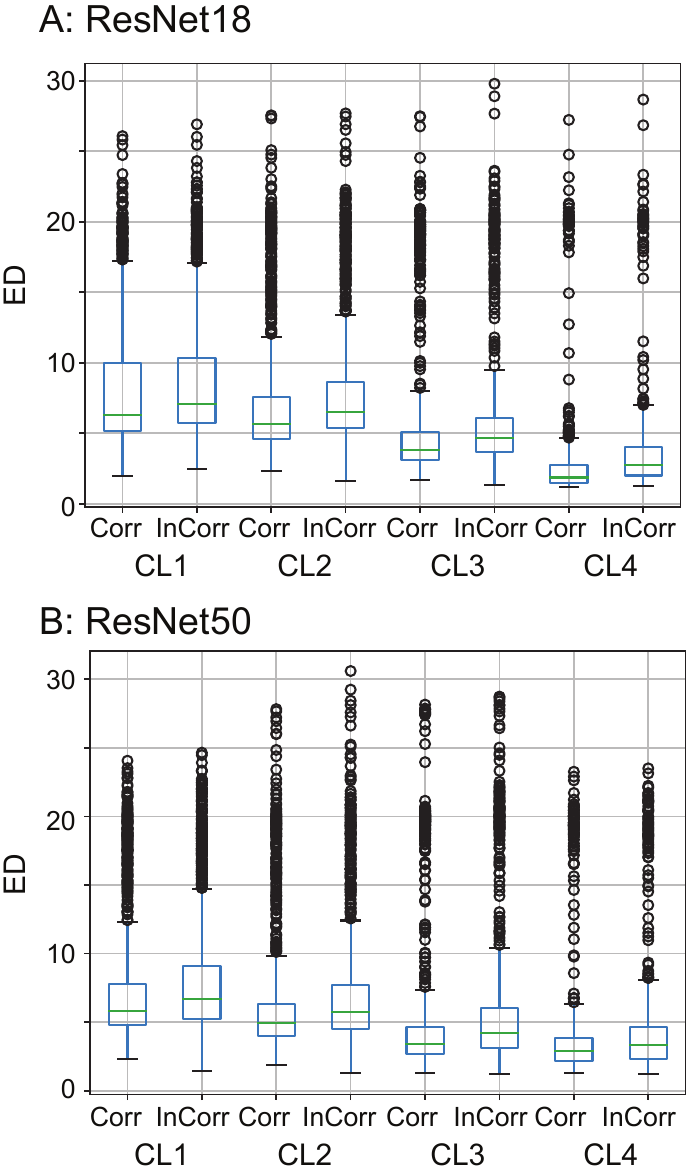}
\caption{EDs between BMUs and the label centers. For each class, we estimate the mean ED over 50 examples from SOM-val; we consider 10 BMUs per input.The label centers are estimated using SOM-tr. $x$-axis denotes the data type. EDs are grouped based on CL and whether ResNets make correct predictions (Corr) or not (InCorr). The open circles in the plots denote the outlier classes. (A), EDs measured from SOMs trained on ResNet18. (B), EDs measured from SOMs trained on ResNet50.}
  \label{supplfig2}

\end{figure*}

\begin{figure*}[htb]
  \centering
  \includegraphics[width=0.6\linewidth]{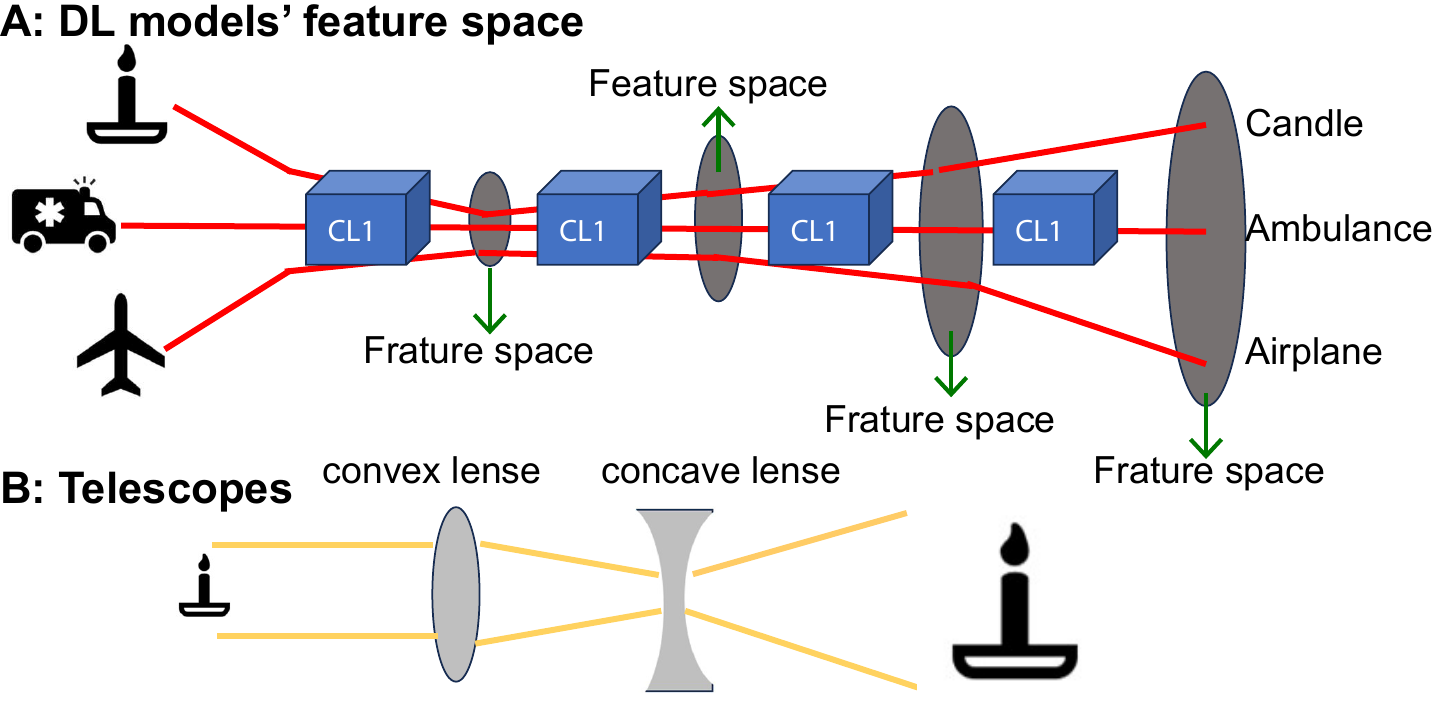}
\caption{Functional Telescope Hypothesis proposes that feature spaces of DL models are condensed and then expanded.}
  \label{FT}
\end{figure*}

\end{document}